\documentclass[runningheads]{llncs}

\usepackage{times}

\usepackage{amssymb}
\usepackage{amsmath}
\usepackage{psfrag}
\usepackage{latexsym}

\usepackage{graphicx,verbatim}

\usepackage{epsfig,algorithm,float,color}

\begin{document}

\mainmatter

\title{A Pseudo-Boolean Solution to the Maximum Quartet Consistency Problem
\thanks{This work is partially supported by
the European Scholarship Program of Microsoft Research.}}
\author{Ant\'{o}nio Morgado \and Jo\~{a}o Marques-Silva}
\institute{School of Electronics and Computer Science, University of Southampton, UK\\
\email{ajrm@soton.ac.uk,jpms@ecs.soton.ac.uk}}
\titlerunning{A PB Solution to the MQC Problem }
\authorrunning{A.~Morgado and J.~Marques-Silva}
\toctitle{}
\tocauthor{}
\maketitle

%%%%%%%%%%%%%%%%%%%%%%%%%%%%%%%%%%%%%%%%%%%%%%%%%%%%%%%%%%%%%%%%%%%%%%%%%%%%%%%
%%

\begin{abstract}
Determining the evolutionary history of a given biological data is an
important task in biological sciences. Given a set of quartet
topologies over a set of taxa, the \emph{Maximum Quartet
  Consistency}~(MQC) problem consists of computing a global phylogeny
that satisfies the maximum number of quartets. A number of solutions
have been proposed for the MQC problem, including Dynamic Programming,
Constraint Programming, and more recently Answer Set Programming
(ASP). ASP is currently the most efficient approach for optimally
solving the MQC problem. This paper proposes encoding the MQC problem
with pseudo-Boolean (PB) constraints. The use of PB allows solving the
MQC problem with efficient PB solvers, and also allows considering
different modeling approaches for the MQC problem. Initial results are
promising, and suggest that PB can be an effective alternative for
solving the MQC problem.
\end{abstract}

\setcounter{footnote}{0}

\section{Introduction}

The amount of existing biological data (DNA and protein sequences) has
increased the need for larger and faster determination of evolutionary
history (or {\em phylogeny}) given a set of taxa (i.e.~a set of related
biological species~\cite{chor-sofsem98}).
Moreover, the availability of data is not always the same for
different taxa. This is known as the data disparity
problem~\cite{wu_wabi05,wu_ifp06}. In recent years, quartet based
methods have received greater attention from the computational biology
community as a way to overcome the data disparity problem.
Quartet-based methods are characterized by first inferring a set
of evolutionary relationships between four taxa, and then from these
relationships assemble a global evolutionary tree. 
Considering only four taxa in the first step to build the evolutionary
relationships, leads to a greater confidence on the relationships
produced. Nevertheless, the relationships obtained may be conflicting
or even missing. The aim of this work is to obtain the evolutionary
tree, under the parsimony assumption, that respects the maximum number
of these relationships on four taxa.

Given a set of quartet topologies over a set of taxa, the
\emph{Maximum Quartet Consistency}~(MQC) problem consists of computing
a global phylogeny that satisfies the maximum number of quartets. A
number of solutions have been proposed for the MQC problem, including
Dynamic Programming, Constraint Programming, and more recently Answer
Set Programming (ASP)~\cite{wu_wabi05,wu_apbc05,wu_tcbb07}. ASP is
currently the most efficient approach for optimally solving the MQC
problem.
This paper develops an encoding for the MQC problem with
pseudo-Boolean (PB) constraints. Initial results are promising, and
suggest that PB can be an effective alternative for solving the MQC
problem.

The paper is organized as follows. The first section introduces both
the MQC problem and the MQI problem. The following section develops a Pseudo Boolean 
Optimization~(PBO) model for the MQC problem and
Section~\ref{sec:optimizations} proposes three optimizations to the PBO
model. Section~\ref{sec:results} shows the experimental results
obtained and Section~\ref{sec:conclusions} presents some conclusions
and points some directions for future research.

% end \input{introduction}
%
% \input{review}

\section{Preliminaries}

\begin{figure}[t]
  \begin{center}
    \begin{tabular}{cp{1cm}cp{1cm}c}     
    \psfrag{x}{$a$}
    \psfrag{y}{$b$}
    \psfrag{z}{$c$}
    \psfrag{w}{$d$}
    \includegraphics[scale=0.4]{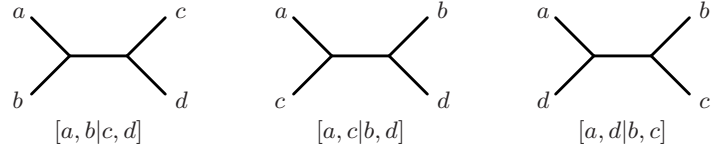}
    & &
    \psfrag{x}{$a$}
    \psfrag{y}{$c$}
    \psfrag{z}{$b$}
    \psfrag{w}{$d$}
    \includegraphics[scale=0.4]{images/xy_zw}
    & &
    \psfrag{x}{$a$}
    \psfrag{y}{$d$}
    \psfrag{z}{$b$}
    \psfrag{w}{$c$}
    \includegraphics[scale=0.4]{images/xy_zw}\\
    $[a,b|c,d]$
    & &
    $[a,c|b,d]$
    & &
    $[a,d|b,c]$
    \end{tabular}
  \end{center}
  \caption{Graphical representation of the quartet topologies $[a,b|c,d]$, $[a,c|b,d]$ and $[a,d|b,c]$.}
  \label{fig:topologies}
\end{figure}

A \emph{phylogeny} is an unrooted tree whose leaves are bijectively mapped to a given set of taxa $S$, where each internal node has degree three.
A \emph{quartet} is a size four subset of $S$.
For each quartet there exist three different possible phylogenies, called \emph{quartet topologies}.
Consider the quartet $\{a,b,c,d\}$, the three possible quartet
topologies will be denoted by $[a,b|c,d]$, $[a,c|b,d]$ and
$[a,d|b,c]$.
Figure~\ref{fig:topologies} gives a graphical representation of the
three possible quartet topologies for the quartet $\{a,b,c,d\}$. For
example, quartet topology $[a,b|c,d]$ means that the path that
connects $a$ and $b$ does not intersect the path connecting $c$ and
$d$.

\begin{figure}[t]
  \begin{center}
    \begin{tabular}{cp{1.5cm}c}
      \psfrag{t1}{$a$}
      \psfrag{t2}{$b$}
      \psfrag{t3}{$c$}
      \psfrag{t4}{$d$}
      \psfrag{t5}{$e$}
      \psfrag{t6}{$f$}
      \psfrag{t7}{$g$}
      \includegraphics[scale=0.6]{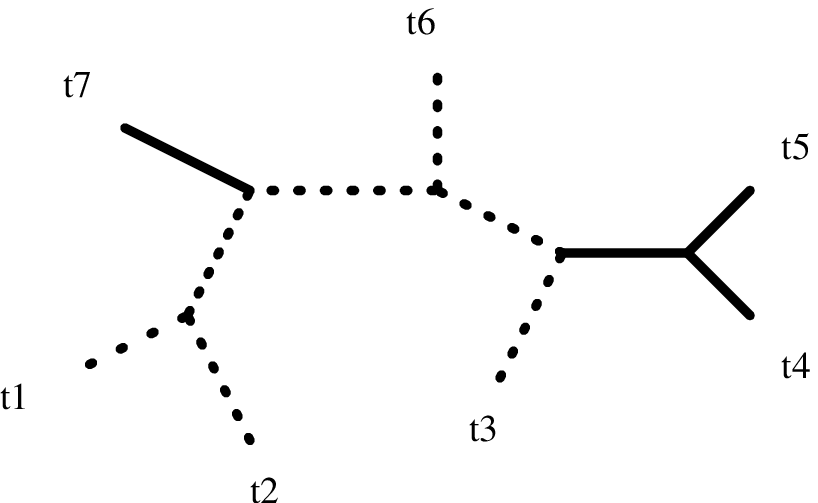}
      & &
      \psfrag{x}{$a$}
      \psfrag{y}{$b$}
      \psfrag{z}{$c$}
      \psfrag{w}{$f$}
      \includegraphics[scale=0.4]{images/xy_zw}
    \end{tabular}
  \end{center}
  \caption{Graphical representation of a phylogeny and of the quartet topology for the  quartet $\{a,b,c,f\}$ derived from the phylogeny.}
  \label{fig:quartetderived}
\end{figure}

Given a phylogeny $T$ on $S$ and a quartet $q=\{a,b,c,d\}$, a quartet
topology $qt$ is said to be the quartet topology of $q$ derived from $T$,
if $qt$ is the topology obtained from $T$, by removing all the
edges and nodes not in the paths connecting the leaves that are
mapped to taxa in $q$.
Figure~\ref{fig:quartetderived} represents a phylogeny, and the
quartet topology derived from the phylogeny for the quartet
$\{a,b,c,f\}$. 
The dotted branches show the path connecting the taxa in the quartet.
Since the path that connects $a$ and $b$ does not intersect the path that connects $c$ and $f$, then the derived quartet topology is $[a,b|c,f]$. 

The set of quartet topologies derived from a phylogeny $T$ is denoted by $Q_T$.
If a quartet topology $q$ is the same as the quartet topology derived from $T$, then $T$ is said to \emph{satisfy} $q$ and $q$ is said to be \emph{consistent} with $T$.
In the example of Figure~\ref{fig:quartetderived}, $[a,b|c,f]$ is consistent with the phylogeny shown, but $[a,c|f,g]$ is not.

Given a set of quartet topologies $Q$ on the set of taxa
$S=\{s_1,\ldots,s_n\}$, if there exists a phylogeny $T$ that satisfies
all the quartet topologies in $Q$, then $Q$ is said \emph{compatible}.
In practice the quartet topologies in $Q$ may be inaccurate or even missing.
If the set $Q$ contains a quartet topology for each possible quartet of $S$, then $Q$ is \emph{complete} otherwise \emph{incomplete}.

The problem of \emph{Maximum Quartet Consistency (MQC)} is the problem where a set of quartet topologies $Q$ on a set of taxa $S=\{s_1,\ldots,s_n\}$ is given, and returns a phylogeny $T$ on $S$, that satisfies the maximum number of quartet topologies of $Q$.

The MQC problem is NP-hard~\cite{berry_esa99} and if $Q$ is complete, then MQC admits a poly\-no\-mial-time approximation scheme~\cite{jiang_focs98}.
If $Q$ is incomplete, then MQC is MAX SNP-hard~\cite{jiang_focs98}.
The dual problem to the MQC is the problem of \emph{Minimum Quartet Inconsistency}~(MQI).
The MQI problem is the problem that given a set of quartet topologies $Q$ (as in the MQC problem), returns a phylogeny that minimizes the number of quartet errors, where the set of quartet errors is the set $Q-Q_T$.
%The MQI problem is NP-hard\cite{berry_esa99,jiang_ja00}, and if $Q$ is complete then MQI admits $O(n^2)$-approximation algorithms\cite{berry_esa99,berry_sda00,jiang_ja00}.
%
The rest of the paper assumes that the set of quartet topologies $Q$
is complete.
In the recent past, different approaches have been reviewed in the
literature for both the MQC and MQI problems. A detailed review is
presented in~\cite{wu_tcbb07}.
\begin{comment}
For exact approaches in 1998 Ben-Dor et al.~\cite{bendor_recomb98} and Pelleg~\cite{pelleg_msc98} have used a dynamic programming approach to solve MQC, and in 2003 Gramm and Niedermeier~\cite{gramm_css03} described an algorithm to solve the MQI but requiring as input the exact number of quartet errors.
In 2004 Wu et al.~\cite{wu_icati04} presented an encoding of the MQC problem to answer set programming, and in 2005 Wu et al.~\cite{wu_wabi05} presented a lookahead branch-and-bound algorithm for the MQC problem.
In 2006, Gang Wu et al.~\cite{wu_ifp06} proposed a polynomial time algorithm for the MQI problem, when the number of quartet errors is $O(n)$.

Heuristic approaches include the quartet puzzling heuristic by Strimmer and Haeseler~\cite{strimmer_mbe96}, the short quartet method by Erdos et al.~\cite{erdos_icalp97}.
In 1998 Ben-Dor et al.~\cite{gramm_css03} used semidefinite programming and in 1999 proposed the quartet cleaning methods~\cite{berry_sda00,berry_esa99,jiang_focs98}.

The focus of this work is on exact approaches for the MQC problem.
\end{comment}

% end \input{review}
%
% \input{pboModel}

\section{Pseudo Boolean Model for the MQC Problem}
\label{sec:pboModel}

This section develops a Pseudo Boolean Optimization(PBO) model for
solving the MQC problem. The idea of the model is to obtain a rooted phylogeny, from which it is possible to construct an unrooted phylogeny~\cite{pelleg_msc98}. 
Similarly to the existing ASP
solution~\cite{wu_tcbb07}, the PBO model encodes the constraints of
representing the rooted phylogeny tree as an ultrametric matrix. 
Moreover, an ultrametric phylogeny satisfies the maximum  number of
quartets topologies of a set $Q$ if and only if the corresponding
ultrametric matrix $M$ satisfies the maximum number of quartets
topologies in $Q$~\cite{wu_tcbb07}. 

%The PBO model encode a rooted phylogeny for the set of taxa given. 
%From rooted phylogenies it is possible to obtain unrooted phylogenies
%and vice-versa~\cite{pelleg_msc98}.

%The idea of the PBO model is to encode an ultrametric matrix $M$,
%from which it is possible to obtain the associated phylogeny tree.

\begin{figure}[t]
  \begin{center}
    \begin{tabular}{cp{1.5cm}c}
      \psfrag{t1}{$a$}
      \psfrag{t2}{$b$}
      \psfrag{t3}{$c$}
      \psfrag{t4}{$d$}
      \psfrag{t5}{$e$}
      \psfrag{t6}{$f$}
      \psfrag{t7}{$g$}
      \includegraphics[scale=0.4]{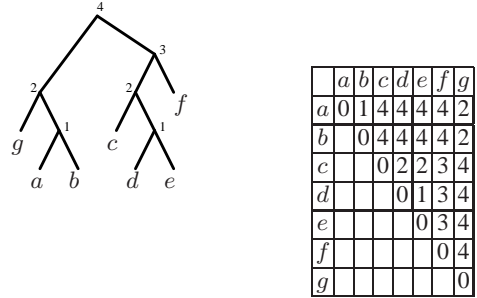}
      & &
      \begin{tabular}{|c|c|c|c|c|c|c|c|}
        \hline
            & $a$ & $b$ & $c$ & $d$ & $e$ & $f$ & $g$\\
        \hline
        $a$ &  0  &  1  &  4  &  4  &  4  &  4  &  2 \\
        \hline
        $b$ &     &  0  &  4  &  4  &  4  &  4  &  2 \\
        \hline
        $c$ &     &     &  0  &  2  &  2  &  3  &  4 \\
        \hline
        $d$ &     &     &     &  0  &  1  &  3  &  4 \\
        \hline
        $e$ &     &     &     &     &  0  &  3  &  4 \\
        \hline
        $f$ &     &     &     &     &     &  0  &  4 \\
        \hline
        $g$ &     &     &     &     &     &     &  0 \\
        \hline
      \end{tabular}
    \end{tabular}
  \end{center}
  \caption{Graphical representation of a rooted phylogeny and the associated ultrametric matrix.}
  \label{fig:rootphyl}
\end{figure}

Consider the set of taxa $S=\{s_1,\ldots,s_n\}$ and a set of quartets
$Q$. An \emph{ultrametric} matrix $M$ is a symmetric square matrix $n
\times n$, where for each $i$ such that $1\leq i \leq n$ then  $M(i,i)
= 0$, for each $i,j$ such that $1\leq i < j \leq n$ then $1\leq
M(i,j) = M(j,i)\leq n$, and for each triple of indices $i,j,k$ such that $1\leq i,j,l\leq n$, there is a tie between the maximum value of $M(i,j)$, $M(i,l)$ and $M(j,l$). 

The values in the ultrametric matrix $M$, represent the lowest common
ancestor in the rooted phylogeny, that is the value of $M(i,j)$
corresponds to the internal node of the phylogeny that is the lowest
common ancestor between taxa $i$ and $j$.
Figure~\ref{fig:rootphyl} presents a rooted phylogeny, where the
internal nodes have been labeled.
The labels correspond to integers in decreasing order from the root to
the leaves.
On the right side of the figure is represented half of the associated
ultrametric matrix.
In~\cite{gusfield_book97} it is explored the relationship between rooted phylogenies and ultrametric matrixes and presents an algorithm to obtain a rooted phylogeny from the associated ultrametric matrix in polynomial time.

It was proven in~\cite{wu_tcbb07} that in order to obtain an optimal
phylogeny, the values of the entries of $M$ can be restricted to $1\leq M(i,j) \leq \lceil \frac{n}{2}\rceil$.
To encode the values of $M(i,j)$ the PBO model introduces a
set of Boolean variables $M_{i,j,k}$ where $1\leq i<j\leq n$ and
$1\leq k\leq\lceil \frac{n}{2}\rceil$. 
$M_{i,j,k}$ has value $1$ iff $M(i,j)=k$, otherwise $M_{i,j,k}$ is $0$.
To ensure that, for each pair $(i,j)$, one and only one of the
variables $M_{i,j,k}$ is selected to be true, the model introduces the following constraint:
\begin{equation}
  \label{eq:encd_unique}
  \sum_{k=1}^{\lceil \frac{n}{2}\rceil} M_{i,j,k} = 1
\end{equation}
The value of each $M(i,j)$ variable is given by $M(i,j) =
\sum_{k=1}^{\lceil\frac{n}{2}\rceil} k\times M_{i,j,k}$.

To ensure that the resulting matrix $M$ is ultrametric, one of the
following three conditions must be satisfied, for each $1\leq
i<j<l\leq n$:
\begin{eqnarray}
M(i,j) = M(i,l) & \wedge & M(i,l) > M(j,l) \mbox{, or} \label{eq:ultram_1}\\
M(i,j) = M(j,l) & \wedge & M(j,l) > M(i,l) \mbox{, or} \label{eq:ultram_2}\\
M(j,l) = M(i,l) & \wedge & M(i,l) > M(i,j) \label{eq:ultram_3}
\end{eqnarray}

The PBO model associates three new Boolean variables $c1_{i,j,l}$,
$c2_{i,j,l}$, $c3_{i,j,l}$ with constraints (\ref{eq:ultram_1}),
(\ref{eq:ultram_2}) and (\ref{eq:ultram_3}), respectively. Each of
the variables $cx_{i,j,l}$ is true iff the associated constraint is
satisfied. 

Constraint (\ref{eq:ultram_1}) is the logical AND of an equality
constraint and a greater than constraint.
In the PBO model each of these constraints is associated with
additional Boolean variables, respectively, $c1_{i,j,l}^1$ and
$c1_{i,j,l}^2$.
$c1_{i,j,l}^1 = 1$ iff $M(i,j) = M(i,l)$, and can be implemented with
a comparator circuit on the unary representation of $M(i,j)$ and
$M(i,l)$, using variables $M_{i,j,k}$ and $M_{i,l,k}$.
$c1_{i,j,l}^2 = 1$ iff $M(i,l) = M(j,l)$, and can also be implemented
with a comparator circuit on the unary representation of $M(i,l)$ and
$M(j,l)$, using variables $M_{i,l,k}$ and $M_{j,l,k}$.
%
\begin{comment}
The actual constraints can be encoded with Boolean comparator circuits.
\begin{eqnarray}
  c1_{i,j,l}^1 &=& Equal(\{M_{i,j,k}|1\leq k\leq \lceil\frac{n}{2}\rceil\},\{M_{i,l,k}|1\leq k\leq \lceil\frac{n}{2}\rceil\})\\
  c1_{i,j,l}^2 &=& Greater(\{M_{i,l,k}|1\leq k\leq \lceil\frac{n}{2}\rceil\},\{M_{j,l,k}|1\leq k\leq \lceil\frac{n}{2}\rceil\})
\end{eqnarray}
\end{comment}
As a result, $c1_{i,j,l}$ is defined as:
\begin{equation}
  c1_{i,j,l} = AND(c1_{i,j,l}^1,c1_{i,j,l}^2)
\end{equation}

Variables $c2_{i,j,l}$ and $c3_{i,j,l}$ are encoded similarly.
Finally to guarantee that one of the conditions (\ref{eq:ultram_1}),
(\ref{eq:ultram_2}) or (\ref{eq:ultram_3}) is satisfied, the PBO
model uses the following constraint:
\begin{equation}
\label{eq:encd_ultram}
c1_{i,j}+c2_{i,j}+c3_{i,j} \geq 1
\end{equation}

As the objective is to compute the phylogeny that maximizes the number
of quartets that can be satisfied, then with each quartet is
associated with a Boolean variable $q_t$, where $1\leq t\leq |Q|$.
$q_t$ will be true if quartet number $t$ is consistent, otherwise
$q_t$ is false.
A quartet $[i,j|l,m]$ is consistent if and only if one of
the following conditions is satisfied~\cite{wu_tcbb07}: 
\begin{eqnarray}
  M(i,l) > M(i,j) & \wedge & M(j,m) > M(i,j) \mbox{, or} \label{eq:cons_1}\\
  M(i,l) > M(l,m) & \wedge & M(j,m) > M(l,m) \label{eq:cons_2}
\end{eqnarray}

Suppose that quartet number $t$ is the quartet $[i,j|l,m]$.
The model associates two new variables to each of the conditions
(\ref{eq:cons_1}) and (\ref{eq:cons_2}). 
Let $d1_{i,j,l,m}$ be associated with condition (\ref{eq:cons_1}) and $d2_{i,j,l,m}$ be associated with condition (\ref{eq:cons_2}).
The associated variable $q_t$ is encoded as a gate OR:
\begin{equation}
  q_t = OR(d1_{i,j,l,m},d2_{i,j,l,m}) 
\end{equation}

Both the conditions (\ref{eq:cons_1}), (\ref{eq:cons_2}) consist of
logical ANDs of two greater than conditions. Thus variable
$d1_{i,j,l,m}$ and $d2_{i,j,l,m}$ are encoded as gates AND in a
analogous way to variables $c1_{i,j,l}$.

The cost function of the PBO model is then to maximize the number of quartets that are consistent, that is:
\begin{equation}
  \label{eq:encd_costfunction}
  \max:\sum_{t=1}^{|Q|} q_t
\end{equation}

% end  \input{pboModel}
%
% \input{optimizations}

\section{Optimizations to the PBO Model}
\label{sec:optimizations}

This section describes three optimizations to the basic PBO model. The
first optimization aims reusing auxiliary variables that serve for
encoding of some of the circuits associated with the PBO model.
The second optimization is related with the Boolean variables used for
representing the value of each entry in the ultrametric matrix.
The third optimization sets the values for some of $M(i,j)$
variables when it is known that $s_i$ and $s_j$ are siblings.

\subsection{First Optimization}
\label{sec:nmqcOptimization}

The objective of the first optimization is to reduce the number of
variables used in the encoding.
The reduction is achieved by exploiting the information provided by
the auxiliary variables used for encoding cardinality constraints.
In order to implement this optimization, sequential
counters~\cite{sinz_cp05} are used.
The uniqueness constraint (\ref{eq:encd_unique}) of the PBO model in
Section~\ref{sec:pboModel} is split into two constraints.
The first constraint deals with the need to have one at least one
variable selected by adding the constraint:
\begin{equation}
\sum_{k=1}^{\lceil \frac{n}{2}\rceil}M_{i,j,k} \ge 1
\end{equation} 

The second constraint is:
\begin{equation}
\sum_{k=1}^{\lceil \frac{n}{2}\rceil}M_{i,j,k} \leq 1
\end{equation}
and is encoded in CNF with a sequencial counter~\cite{sinz_cp05}.
This sequential counter introduces variables $s_{k,1}$.
These variables have the property that if $M_{i,j,a}=1$ then for $1\leq
k<a$ all variables have $s_{k,1}=0$ and for $a\leq k\leq \lceil\frac{n}{2}\rceil$ then
$s_{k,1}=1$.
The property enables the encoding of $M(i,j)< M(l,m)$
by considering the associated variables $s_{k,1}$ of $M(i,j)$ and of
$M(l,m)$.
In order to better understand, let the variables $s_{k,1}$ associated
to the sequential counter of $M(i,j)$ be denoted by $s_{k}^{i,j}$.
The objective is to encode that $M(i,j) < M(l,m)$ by re-using the
variables $s_{k}^{i,j}$ and $s_{k}^{l,m}$.
Using the above property, this can be done by searching for the $k$ where
$s_k^{i,j}=1$ and $s_k^{l,m}=0$, which can be encoded in a variable
$e_k^{(i,j)(l,m)}$ as a gate AND:
\begin{equation}
e_k^{(i,j)(l,m)}=AND(s_k^{i,j},NOT(s_k^{l,m}))
\end{equation}
Then variable $LT_{i,j,l,m}$ encodes that $M(i,j)< M(l,m)$ by a gate
OR:
\begin{equation}
LT_{i,j,l,m}=OR(e_k^{(i,j)(l,m)}: 1\leq k\leq\lceil\frac{n}{2}\rceil)
\end{equation}

For this optimization, all the other constraints of the PBO model of
Section~\ref{sec:pboModel} are maintained, but making use of the
variables $LT_{i,j,l,m}$ as appropriate.

\subsection{Second Optimization}
\label{sec:logOptimization}

For the PBO model described in Section~\ref{sec:pboModel}, for each
pair of taxa $(i,j)$, the values of the variables $M(i,j)$ are
encoded through selection variables $M_{i,j,k}$ where $1\leq k\leq
\lceil\frac{n}{2}\rceil$. 

The first optimization described here replaces the encoding of the
selection variables.
Variables $M_{i,j,k}$ are still going to be used to encode $M(i,j)$,
but here $M_{i,j,k}$ represents the $k-$th bit of the binary
representation of $M(i,j)$.
Now $k$ is limited by $0\leq k\leq
\lfloor\log_2(\lceil\frac{n}{2}\rceil)\rfloor$.
With this encoding $M(i,j)$ can be obtained by $M(i,j) = \sum_{k=0}^{\lfloor\log_2(\lceil\frac{n}{2}\rceil)\rfloor}2^k\times M_{i,j,k}$.
Moreover, the constraints used in the encoding need to be
modified. The constraints in Equation~(\ref{eq:encd_unique}) that
encode the uniqueness of the selection variables are no longer used. All the
other constraints are maintained, but with the new limit for variable
$k$.
Instead of the uniqueness constraints, this optimization requires that
the encoded variables $M(i,j)$ are restricted to
$\{1,\ldots,\lceil\frac{n}{2}\rceil\}$, that is $ 1\leq M(i,j)$ and
$M(i,j)\leq \lceil\frac{n}{2}\rceil$. 
The first part is obtained by adding the constraint:
\begin{equation}
  \sum_{k=0}^{\lfloor\log_2(\lceil\frac{n}{2}\rceil)\rfloor} M_{i,j,k} \geq 1
\end{equation}
For the second part, a new Boolean variable $ltb_{i,j}$ is introduced,
that captures the condition that $M(i,j)$ is not larger than
$\lceil\frac{n}{2}\rceil$. The variables $M_{i,j,k}$ are used to
representing this constraint as a comparator circuit.

\begin{comment}
This variable encodes the satisfaction that $M(i,j)$ is lower than a bound.
To encode it on the PBO model, a comparator circuit is used, where one of the input bits are the bits to encode $\lceil\frac{n}{2}\rceil$ into binary, that is:
\begin{equation}
  ltb_{i,j} = Lower(\{M_{i,j,k}\},\{b_k| \sum_{k=0}^{\lfloor\log_2(\lceil\frac{n}{2}\rceil)\rfloor} 2^k\times b_k= \lceil\frac{n}{2}\rceil\})
\end{equation} 
\end{comment}

In order to ensure that $ltb_{i,j}$ is true, the following constraint is added to the model:
\begin{equation}
  ltb_{i,j} \geq 1
\end{equation}

\subsection{Third Optimization} 
\label{sec:siblingsOptimization}

The optimization described in this section
follows~\cite{wu_wabi05,wu_apbc05,wu_tcbb07}.
The objective of this optimization is to previously determine the value of some
variables, namely when a pair of taxa is know to be siblings. 
The optimization can be used independently of the model (or
optimization) used.

Let $S=\{s_1,\ldots,s_n\}$ be a set of taxa and $Q$ be a complete set of quartets .
A \emph{Bipartition} of $S$ is a pair $(X,Y)$ of nonempty subsets of $S$, such that $S=X\cup Y$ and $X\cap Y=\emptyset$.
Consider a bipartition $(X,Y)$ of $S$, such that $|X|\geq 2$  and $|Y|\geq 2$, let $Q_{(X,Y)}$ be defined as $Q_{(X,Y)}=\{[x_1,x_2|y_1,y_2]:x_i\in X \wedge y_i\in Y \mbox{ for }i\in\{1,2\}\}$.
Suppose that three taxa from $Y$ are fixed and also that $|X|=l$.
An \emph{$l$-subset} with respect to $(X,Y)$ is the set of $l$ quartets from $Q$ that contain the three fixed taxa from $Y$ and one taxa from $X$.
There are a total of $\binom{n-l}{3}$ of $l$-subsets.

An $l$-subset is said to be \emph{exchangeable} on X, if by ignoring the difference of the taxa from $X$ on the quartets in the $l$-subset, it produces a unique quartet topology, otherwise the $l$-subset is said to be \emph{nonexchangeable}. 
In the case where $l=2$, then both taxa in $X$ are said to be siblings and the following corollary holds:
\begin{proposition}[Corollary 2.5 from~\cite{wu_tcbb07}] % \cite{wu_trcsua05}
Let $S=\{s_1,\ldots,s_n\}$ be a set of taxa, $Q$ be a complete set of  quartets on taxa $S$.
For the pair of taxa $(s_i,s_j)$ from $S$, let $p_1=|Q_{(\{s_i,s_j\},Y)}-Q|$, $p_2$ be the number of nonexchangeable pairs on $\{s_i,s_j\}$.
If $2p_1+p_2\leq n-3$ then $s_i$, $s_J$ are siblings in an optimal phylogeny.
\end{proposition}
 
In the optimization described in this section, for every pair of
taxa, the condition of the corollary is tested. 
When the condition is true, for example for taxa $i$ and $j$, then the PBO model is augmented with the following constraints:
\begin{eqnarray}
M_{i,j,1} &\geq&1 \label{eq:pbo_opts_1}\\
\nonumber\\
-1\times M_{i,j,k}&\geq&0\quad
\mbox{, $k\in \{2,\dots,upperLimit\}$}\label{eq:pbo_opts_k}
\end{eqnarray}

The $upperLimit$ in Equation~(\ref{eq:pbo_opts_k}) is dependent on the encoding of variable $M_{i,j,k}$ (either as described in Section~\ref{sec:pboModel} or as described in Section~\ref{sec:logOptimization}).

% end \input{optimizations}
%
% \input{results}

\section{Experimental Results}
\label{sec:results}

This section presents experimental results comparing the PBO model
proposed in Section~\ref{sec:pboModel} and the ASP model described
in~\cite{wu_tcbb07}.
The instances considered were obtained from~\cite{wu_tcbb07}. 
These instances correspond to quartet topologies derived from random
generated trees with a percentage of quartet topologies randomly altered.
The percentage of altered quartet topologies introduces errors in the
quartet topologies.
Higher percentage of altered quartet topologies means a higher
possibility of errors in the quartet topologies of the instance.

In the experiments four models were considered, three obtained from the
PBO formulation and one from the ASP formulation.
The first PBO model considers the first optimization described in
Section~\ref{sec:nmqcOptimization} and will be referred as \emph{PBO+fst}.
The second PBO model includes both the optimizations of
Section~\ref{sec:logOptimization} and
Section~\ref{sec:siblingsOptimization}.
This second model will be referred as \emph{PBO+(scd+trd)}. 
The last PBO model, called \emph{PBO+trd}, includes only the third proposed optimization (Section~\ref{sec:siblingsOptimization}).
In all the PBO models an encoder was implemented that receives as
input the quartet topologies and returns as output a file in PB format.
The generated file was then given as input to the PBO solver.
For all experiments the PBO solver used was
\emph{minisat+}~\cite{een_jsat06}.

The fourth model is the ASP model described in~\cite{wu_tcbb07}.
The {\tt phy} program, that encodes the quartet topologies into answer
set programming, was obtained from~\cite{wu_tcbb07}. The instances
were given to {\tt phy}, and for each, the parameters given were the
number of taxa involved and the maximum number of quartet errors known
in the instance. This last parameter was set as the number of quartet
topologies in the instance. After obtaining the encoded instance, the
encoded file was given to the ASP-solver SModels~\cite{simons_smodels} 
%~\footnote{http://www.tcs.hut.fi/Software/smodels/}.
SModels was configured to obtain all the stable models in order to
maximize the number of quartets satisfied.

\begin{table}[t]
  \begin{center}
    \begin{tabular}{|c|c|c|c|c|c|c|}
      \hline
      & \multicolumn{3}{c|}{N. Variables} &
      \multicolumn{3}{c|}{N. Constraints}\\
      \hline
      \% Altered & PBO+fst & PBO+(scd+trd)& PBO+trd & PBO+fst & PBO+(scd+trd)& PBO+trd\\
      \hline
      01 & 5760 & 4514.4 & 6276.6 & 19890 & 16238.8 & 24464   \\
      05 & 5760 & 4537.2 & 6310.8 & 19890 & 16301.5 & 24568.5 \\
      10 & 5760 & 4566.4 & 6354.4 & 19890 & 16385.2 & 24708   \\ 
      15 & 5760 & 4587.6 & 6386.4 & 19890 & 16448.8 & 24814   \\
      20 & 5760 & 4611.2 & 6421.8 & 19890 & 16519.6 & 24932   \\
      25 & 5760 & 4628.4 & 6447.6 & 19890 & 16571.2 & 25018   \\
      30 & 5760 & 4648.4 & 6477.6 & 19890 & 16631.2 & 25118   \\
      \hline
    \end{tabular}
    \caption{Average number of variables and number of constraints for instances with 10 taxa.}
    \label{tab:res_nvar_nconst_10tx}
  \end{center}
\end{table}

\begin{table}[t]
  \begin{center}
    \begin{tabular}{|c|c|c|c|c|}
      \hline
      & \multicolumn{4}{c|}{CPU Time} \\
      \hline
      \% Altered & phy+SModels & PBO+fst & PBO+(scd+trd) & PBO+trd\\
      \hline
      01 & 0.0464  & 0.7696   & 0.4704    & 0.7316   \\
      05 & 0.3048  & 2.2673   & 1.686     & 7.0885   \\
      10 & 1.3264  & 5.7819   & 5.8872    & 28.8291  \\
      15 & 2.4324  & 12.7119  & 11.78235  & 52.6487  \\
      20 & 9.0915  & 32.2536  & 17.78277  & 68.77968 \\
      25 & 28.4901 & 60.7041  & 28.0254   & 117.6832 \\
      30 & 65.4176 & 121.3564 & 52.75086  & 239.2057 \\
      \hline
    \end{tabular}
    \caption{Average CPU time in seconds for instances with 10 taxa.}
    \label{tab:res_10tx}
  \end{center}
\end{table}

The results were obtained on an Intel Xeon 5160, 3GHz server, with
4~GB of RAM.
The results comparing the average number of variables and number of
constraints between the three PBO models is shown in
Table~\ref{tab:res_nvar_nconst_10tx}.
As can be seen from the table the model that requires more variables
and more constraints is the PBO+trd model, whereas, the model that
requires less variables and less constraints is the PBO+(scd+trd).

Table~\ref{tab:res_10tx} compares the average CPU times on the
instances considered for all the PBO models and the phy+Smodes model.

A few conclusions can be drawn from the results.
First comparing the \emph{PBO+fst} and the basic \emph{PBO+trd} model.
The sharing of auxiliary variables introduced by the first
optimization is an important aspect in this problem. 
This optimization reduces the number of variables used by the encoding as well as the
number of constraints.
This reduction leads to lower CPU time spent by the PBO-solver.
Nevertheless, model \emph{PBO+(scd+trd)} reduces even further the
model by considering the selection variables as bits of the binary
representation of values in $M$.
Again, it can be seen from Table~\ref{tab:res_10tx}, that the
reduction on the number of variables and constraints used by the
encoding resulted in lower CPU times spent by the PBO-solver,
where the model \emph{PBO+(scd+trd)} is on average approximately 4
times faster than the \emph{PBO+trd} and 1.6 times faster than \emph{PBO+fst}.

Comparing the best of our PBO models (\emph{PBO+(scd+trd)}) with the
ASP model, the ASP model is more effective when the percentage of
modified quartets is small, but the \emph{PBO+(scd+trd)} model becomes
more when the percentage of modified quartets increases.

% end \input{results}
%
% \input{conclusion}

\section{Conclusions}
\label{sec:conclusions}

This paper proposes a first attempt at solving the MQC problem with 
PBO. The new PBO model is compared with a recent solution based on
ASP~\cite{wu_tcbb07}, which is currently the most efficient for the
MQC problem.
Despite the number of the taxa considered being modest, the results
show that the PBO model can be beneficial when the number of expected
quartet errors is high. The PBO model is still recent, and additional
modeling insights and corresponding performance improvements are to be
expected in the near future.

Future research will involve developing optimizations to the PBO
model. For example, by encoding with PB constraints some of the
optimizations proposed in the literature for the MQC problem. 
Furthermore, experiments will consider larger sets of taxa as well as
real world data. 
% end \input{conclusion}

%%%%%%%%%%%%%%%%%%%%%%%%%%%%%%%%%%%%%%%%%%%%%%%%%%%%%%%%%%%%%%%%%%%%%%%%%%%%%%%
%%

\bibliographystyle{abbrv}
%\bibliography{bibwcb08}

\end{document}